\RequirePackage{fix-cm}
\documentclass{svjour3}                     
\smartqed  
\usepackage{graphicx}
\usepackage{multicol}
\usepackage{subcaption}
\usepackage{soul}
\usepackage[table]{xcolor}
\usepackage{lineno,hyperref}
\usepackage{multirow}
\usepackage{tabularx}
\usepackage{arydshln}
\usepackage{amssymb}
\usepackage[sort&compress,numbers]{natbib}
\usepackage{tikz}
\usetikzlibrary{patterns}
\usepackage{pgfplots}
\usepackage{pgfplots,lipsum}
\usepackage{pdflscape}
\usepackage[margin=1in]{geometry}
\begin{document}

\title{Multimodal price prediction
}


\author{Aidin Zehtab-Salmasi    \and
        Ali-Reza Feizi-Derakhshi$^*$    \and
        Narjes Nikzad-Khasmakhi \and
        Meysam Asgari-Chenaghlu  \and
        Saeideh Nabipour
}


\institute{Aidin Zehtab-Salmasi \at
              Computerized Intelligence Systems Laboratory, Department of Computer Engineering, University of Tabriz, Tabriz, IRAN. \\
              \email{a.zehtab97@ms.tabrizu.ac.ir}        \\
              Orcid: 0000-0001-7410-0299
            \and
            Ali-Reza Feizi-Derakhshi    \at
                Computerized Intelligence Systems Laboratory, Department of Computer Engineering, University of Tabriz, Tabriz, IRAN. \\
                \email{derakhshi96@ms.tabrizu.ac.ir}    \\
                Orcid: 0000-0003-3036-1651
          \and
            Narjes Nikzad-Khasmakhi \at
                Computerized Intelligence Systems Laboratory, Department of Computer Engineering, University of Tabriz, Tabriz, IRAN. \\
                \email{n.nikzad@tabrizu.ac.ir}  \\
                Orcid: 0000-0003-3536-1343
            \and
            Meysam Asgari-Chenaghlu \at
              Computerized Intelligence Systems Laboratory, Department of Computer Engineering, University of Tabriz, Tabriz, IRAN. \\
              \email{m.asgari@tabrizu.ac.ir}    \\
                Orcid: 0000-0002-7892-9675
            \and
            Saeideh Nabipour    \at
                Department Computer and Electrical Engineering, University of Mohaghegh Ardabili, Ardabil, IRAN.    \\
                Orcid: 0000-0002-3622-1513  \\
                \email{saeideh.nabipour@gmail.com}
}

\date{Received: date / Accepted: date}

\maketitle

\begin{abstract}

Price prediction is one of the examples related to forecasting tasks and is a project based on data science. Price prediction analyzes data and predicts the cost of new products. The goal of this research is to achieve an arrangement to predict the price of a cellphone based on its specifications.  So, five deep learning models are proposed to predict the price range of a cellphone, one unimodal and four multimodal approaches. The multimodal methods predict the prices based on the graphical and non-graphical features of cellphones that have an important effect on their valorizations. Also, to evaluate the efficiency of the proposed methods, a cellphone dataset has been gathered from GSMArena. The experimental results show 88.3\% F1-score, which confirms that multimodal learning leads to more accurate predictions than state-of-the-art techniques.

\end{abstract}


\section{Introduction}\label{Sec:Introduction}
With increasing the volume of unstructured data created by IoT devices, humans and social media platforms, data science has become a great analytic science to extract useful and valuable information from this data \cite{Tien2017}.  On the other hand, data mining techniques such as Decision Tree, Neural Network and K-means have accelerated the data science growth to discover patterns  and  extract  knowledge  from a large amount of data stored in different sources such as web \cite{olson2007introduction, sohrabi2012framework, jian2013parallel}. Moreover, data science and machine learning work in a perfect harmony to process data and learn from it for specific tasks such as making predictions. This collaboration helps both businesses and customers. For example, the stock market prediction is a business activity that most people are interested in. Prediction and analysis of stock market data have played a crucial role in today’s economy \cite{Hiransha2018}. From customers' prespective, stock market prediction has significant effects on their future choices such as the decision processes of traders and investors to buy or sell a stock.

The price prediction is one of the examples related to forecasting tasks and
is a project based on data science. The price prediction  analyzes data and predicts the cost of new products. Examples of price prediction include stock prediction \cite{Parmar2018,TeixeiraZavadzkidePauli2020,Singh2017,Karimuzzaman2021}, oil price \cite{SenGupta2021}, electricity price forecasting \cite{Vilar2018}, ticket pricing \cite{Abdella2019,Lantseva2015}, cryptocurrency forecasting \cite{Azari2019,Kadiroglu2019,Phaladisailoed2018,Lahmiri2019,Ji2019}, product price prediction such as house \cite{park2015using} and smartphone \cite{Chandrashekhara2019}. Because of IoT usage of cellphones \cite{Tien2017}, cellphone price prediction is selected as the benchmark of this study.  

Price prediction can be seen as a regression or classification task. For the regression task, the actual price is predicted and for the classification task, the price range in the form of classes is predicted. The accuracy of predicted price can be changed based on the number of classes.  On the other hand, the price prediction task itself, regardless of being regression or classification, is achieved using the features of the products. These features can be in any modality and form. For example, a product such as a cellphone has features in both numerical and image forms. Combination of various and different modalities requires a more robust model that can generalize its learning for samples with different aspects. In the case of cellphone price prediction, the RGB image data that is in form of a $M\times N\times 3$ matrix is combined with features such as 3G network support that is a boolean feature.

Some previous works such as \cite{Nasser2019,Asim2018,Chandrashekhara2019} proposed various methods for the problem of cellphone price prediction. Using a variety of techniques such as machine learning and deep learning requires a huge number of data for training. The lack of such data prohibits researchers to use these approaches on cellphone price prediction. Hence, in this paper, a real cellphone dataset, called CD18,  for applying learning-based approaches to cellphone price prediction has been provided. Moreover, in the current study, five novel models are proposed that aim to predict prices more accurately. One of the proposed models is unimodal and others are based on multimodal learning. Multimodal prediction and learning are also popular in other fields of AI \cite{asgari2020multimodal,nikzad2020berters}.
Using different sentiment analysis and text classification methods \cite{minaee2020deep,vasfisisi2013text} can be investigated even in the text related to user comments. In this paper, the multimodal approach has two data types: graphical features and non-graphical features. While images of the products create graphical features, all other features are defined as non-graphical features. Briefly, the combination of the collected dataset with the proposed multimodal learning approaches provides an insightful view of the cellphone price prediction problem.

\begin{figure*}
	\centering
	\includegraphics[width=\textwidth]{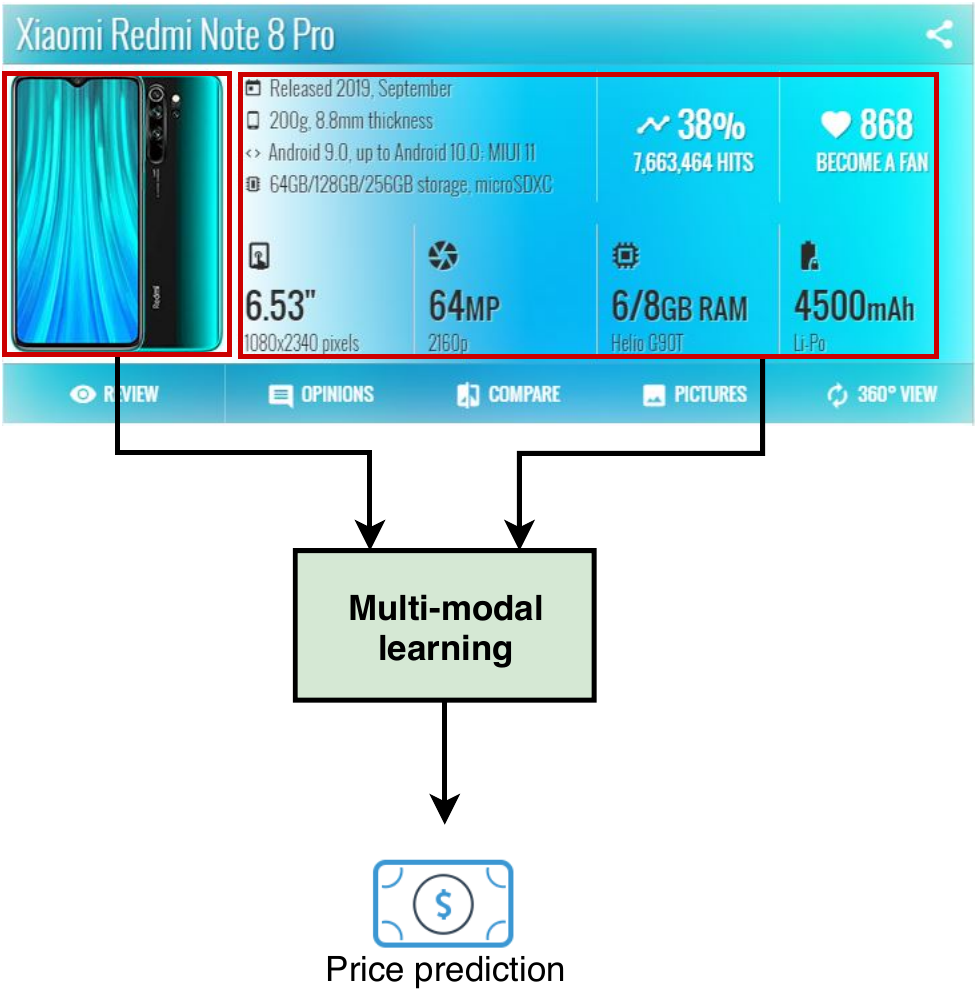}
	\caption{An overview of the idea expressed. Also features that GSMArena offered as essence and collected in this dataset. Example "Xiaomi Redmi Note 8 Pro"}
	\label{Fig:Overal}
\end{figure*}

The rest of this study is organized in the following manner: the second section contains a literature review of mobile price prediction. Section \ref{Sec:Method} presents the proposed methods. The experimental results are given and discussed in section \ref{Sec:Experiments}. Some conclusions are drawn in the final section, and the areas for further researches are identified, as well.

\section{Related study} \label{Sec:Related-study}
While price prediction has a long history in many fields, there are not many works in the product price prediction especially cellphone price prediction that is the main objective of this study. In the following paragraphs, an overview of studies related to the cellphone price prediction is provided. 

Asim  and  Khan \cite{Asim2018} treated the task of cellphone price prediction in the form of classification problem. In this study, prices were classified into four classes (very economical ($<$ 150), economical (151 - 300), expensive (301 - 450), very expensive ($>$ 450)). Two classifiers decision tree (J48) and Naive Bayes were used to obtain the classes. Also, the authors collected a dataset which includes ten features of $134$ cellphones from GSMArena website. In the gathered data, two features brand including categorical and numerical have been used. The memory card slot is categorical, and the rest of the features are numerical.

Also, Nasser and Al-Shawwa \cite{Nasser2019} observed the price prediction as a classification process and used an artificial neural network (ANN) that has three layers. Authors reported their results over Sharma  \cite{Sharma} dataset and considered twenty features of mobile devices and four classes of prices as inputs and outputs, accordingly.  

Unlike the previous methods, the study \cite{Chandrashekhara2019} utilized three regression models for the cellphone price prediction task. These three regression methods include Support vector regression (SVR), Multiple linear regression, and Back-propagation neural network. Also, in this paper, $13$ features and prices for $262$ cellphones were used to evaluate the proposed approaches.

Moreover, in \cite{Subhiksha2020}, the authors investigated the efficiency of three regression models including SVM, Logistic regression, and Random forest over Sharma dataset to forecast the prices.  They reported that SVM and logistic regression have reached the best accuracy ($81\%$). 

Furthermore, Pipalia and Bhadja \cite{Pipalia2020} studied different classifiers to predict cellphones price. Authors deployed five classifiers namely Logistic Regression, K-Nearest Neighbours (KNN), Decision Tree, SVM, and Gradient Boosting to classify the range of mobile price. The benchmark of their research was Sharma dataset. The obtained results of this paper show that Gradient Boosting is the best classifier that reaches $90\%$ accuracy. 

Due to the previous researches, it can be said that there are not multimodal learning and deep learning techniques in cellphone price prediction. So, in this paper, a multimodal learning technique has been employed to combine graphical and non-graphical features. We take the advantages of deep learning structures such as Inception-V3 and convolutional neural network to preserve both features. In this way, each cellphone is presented by a low-dimensional vector that is the concatenation of two features. Then, this vector is fed into a classifier to predict its price. 
\section{Proposed Method} \label{Sec:Method}
This paper aims to propose a novel multimodal approach for predicting the prices of cellphones. As described before, we treat the price prediction task as a classification problem. To achieve it, five methods have been proposed that one of them is a unimodal model, and the rest of the methods are multimodal approaches. The unimodal modal accepts the numerical features as inputs, while the multimodal approaches add the images of cellphones as another modality.  In both unimodal and multimodal approaches, a range of prices for each cellphone are predicted. Each price range is defined as a class of classification task. The important item that makes difference among the proposed multimodal approaches is the way of obtaining features from non-graphical and graphical inputs. Inception-V3 and the convolutional neural networks are deployed to capture image features. On the other hand, non-graphical features are extracted using two different neural networks which consist of three dense layers and convolutional neural networks, respectively. In the next step, both abstracted features are concatenated to create a simple representation for each cellphone sample. Finally, this single feature vector is fed into a classifier to predict and classify the range of cellphone prices. A neural classifier is proposed here which can be exchanged with any other one. In the following paragraphs, the proposed methods are explained in detail. 

\subsection{Model 1: Unimodal} \label{Subsec:model_1}
\begin{figure*}
    \centering
    \includegraphics[width=0.4\textwidth]{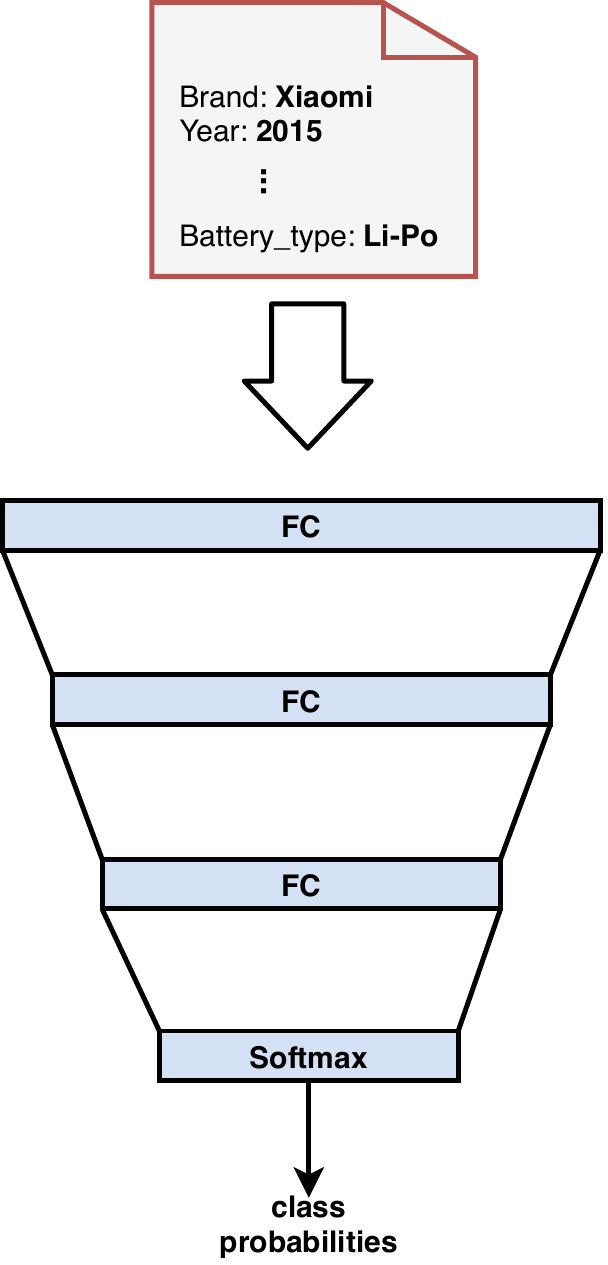}
    \caption{The neural network architecture of the proposed unimodal method.}
    \label{fig:unimodal_model}
\end{figure*}

The proposed unimodal method uses non-graphical features of cellphones to predict and classify prices. As shown in Figure \ref{fig:unimodal_model}, this unimodal approach proposes the utilization of the neural network architecture that consists of three fully connected layers and a softmax layer that transforms the prices into price ranges. The input and output of this approach are the numerical features and a price range as a class of classification, respectively.

\subsection{Model 2: Inception-based image feature extraction}\label{Subsec:model_2}
\begin{figure*}
	\centering
	\includegraphics[width=\textwidth]{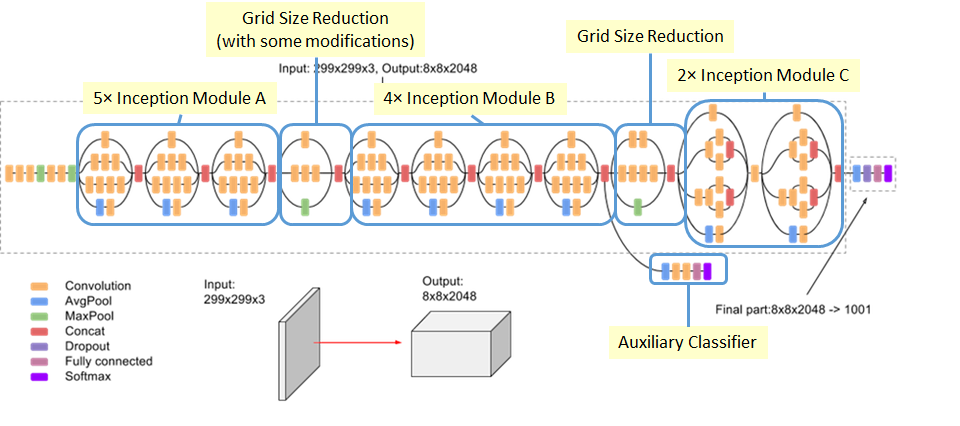}
	\caption{Architecture of Inception-v3  \cite{MediumInception}.}
	\label{Fig:Inception-v3}
\end{figure*}

The second model is a multimodal approach that takes the image and non-graphical features of cellphones as inputs and forecasts the price range. The non-graphical features are extracted by a neural network that consists of three fully connected layers used in the unimodal technique. Moreover, there are various methods to model an image or extract features from it. The most popular one is the convolutional neural network (CNN) \cite{LeCun1989}. This neural network is the foundation of many models, such as LeNet \cite{LeCun1998}, AlexNet \cite{Krizhevsky2017}, GoogleNet \cite{Szegedy2015}, and Inception \cite{Szegedy2016,Szegedy2017}. 
In this study, Inception-V3 model has been used to learn graphical features. For more comprehensive understanding, Figure \ref{Fig:Inception-v3} illustrates the architecture of Inception-V3.  To obtain the image features, the output of the second "Grid Size Reduction" (Figure \ref{Fig:Inception-v3}) of Inception-V3 (pre-trained model) has been selected as image embeddings.

\begin{figure*}
    \centering
    \includegraphics[width=\textwidth]{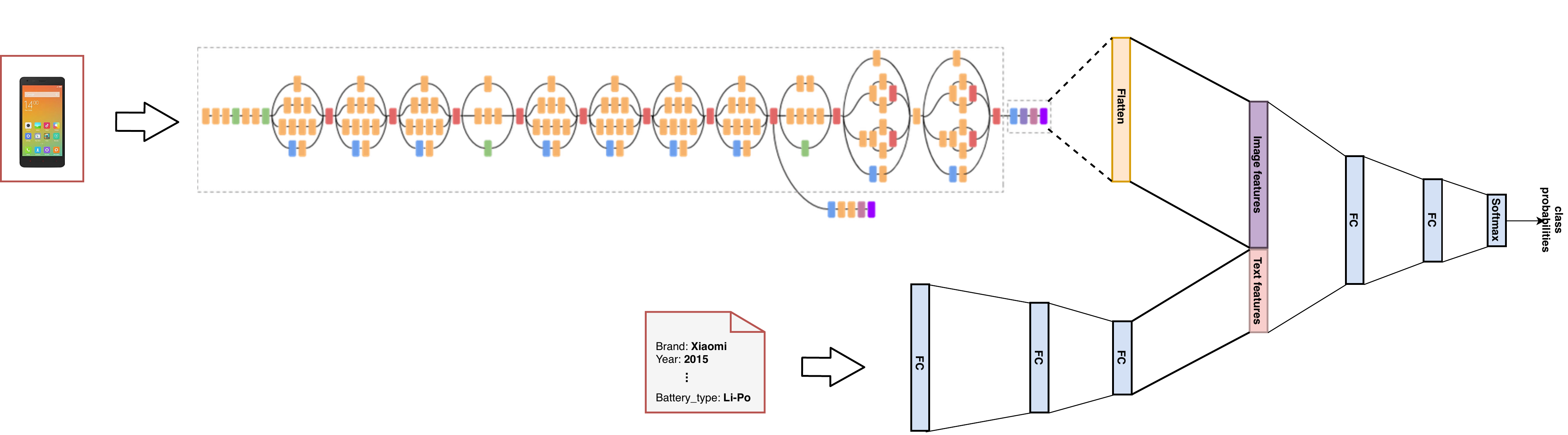}
    \caption{Architecture of the proposed Inception-based method (Model 2).}
    \label{Fig:Inception_model}
\end{figure*}

At the end, both of these embeddings (non-graphical and graphical features) are concatenated and fed into a classifier. A neural network classifier is proposed here which can be replaced with any other classifier such as Logistic Regression. Figure \ref{Fig:Inception_model} demonstrates the structure of the second proposed model.

\subsection{Model 3: Convolutional image feature extraction and dense concatenating}\label{Subsec:model_3}
\begin{figure*}
    \centering
    \includegraphics[width=\textwidth]{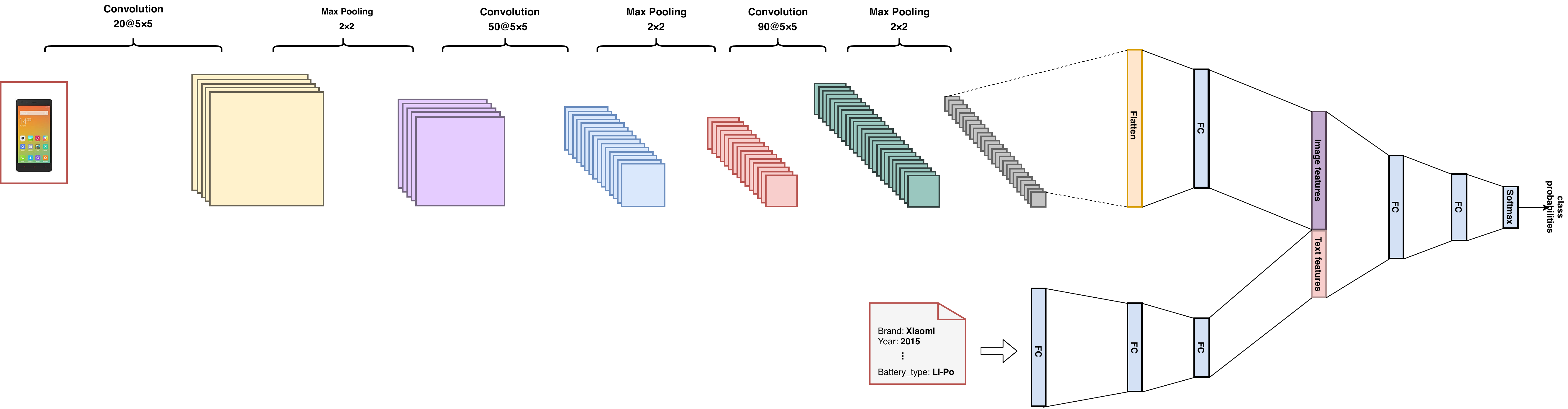}
    \caption{The neural network architecture of the third model.}
    \label{Fig:model_3}
\end{figure*}

Figure \ref{Fig:model_3} presents the third proposed approach. It can be inferred that the embedding of the non-graphical inputs are created by a similar process done in the second proposed model and the unimodal method. Moreover, since the convolutional neural networks are one of the most popular and practical deep learning methods to extract features from images, these networks have been used as the image feature extraction method for the rest of the proposed approaches.  The CNN architecture has three convolutional layers, three maxpooling layers and one fully connected layer. Then, for each cellphone, the embeddings are extracted from non-graphical and image features are integrated into a single representation through the concatenation process. Finally, this single representation is fed into a neural network that includes two dense layers to predict prices.

\begin{figure*}
    \subfloat[First layer of convolution.]{{\includegraphics[width=0.3\linewidth]{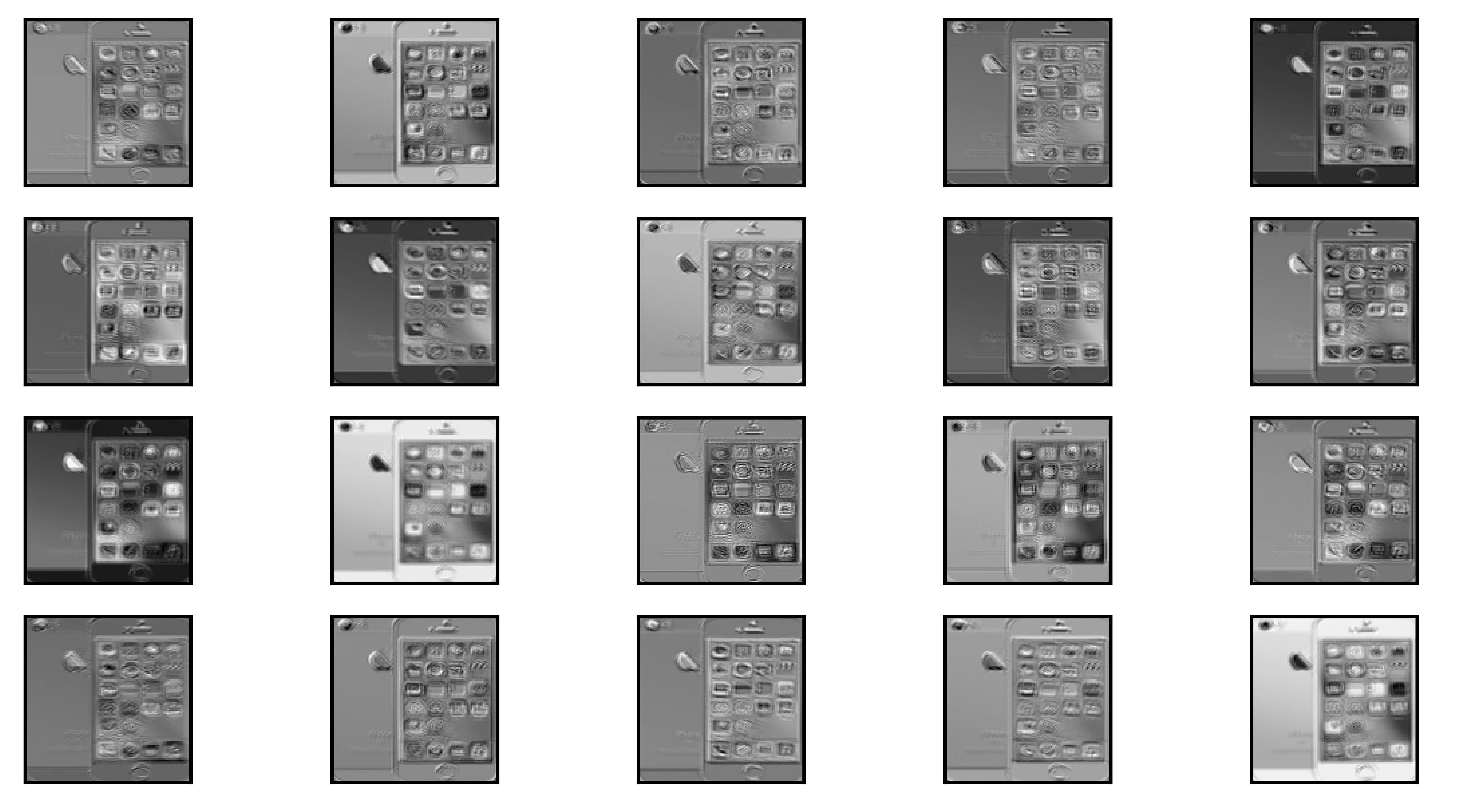}}}%
    \hfill
    \subfloat[Second layer of convolution.]{{\includegraphics[width=0.3\linewidth]{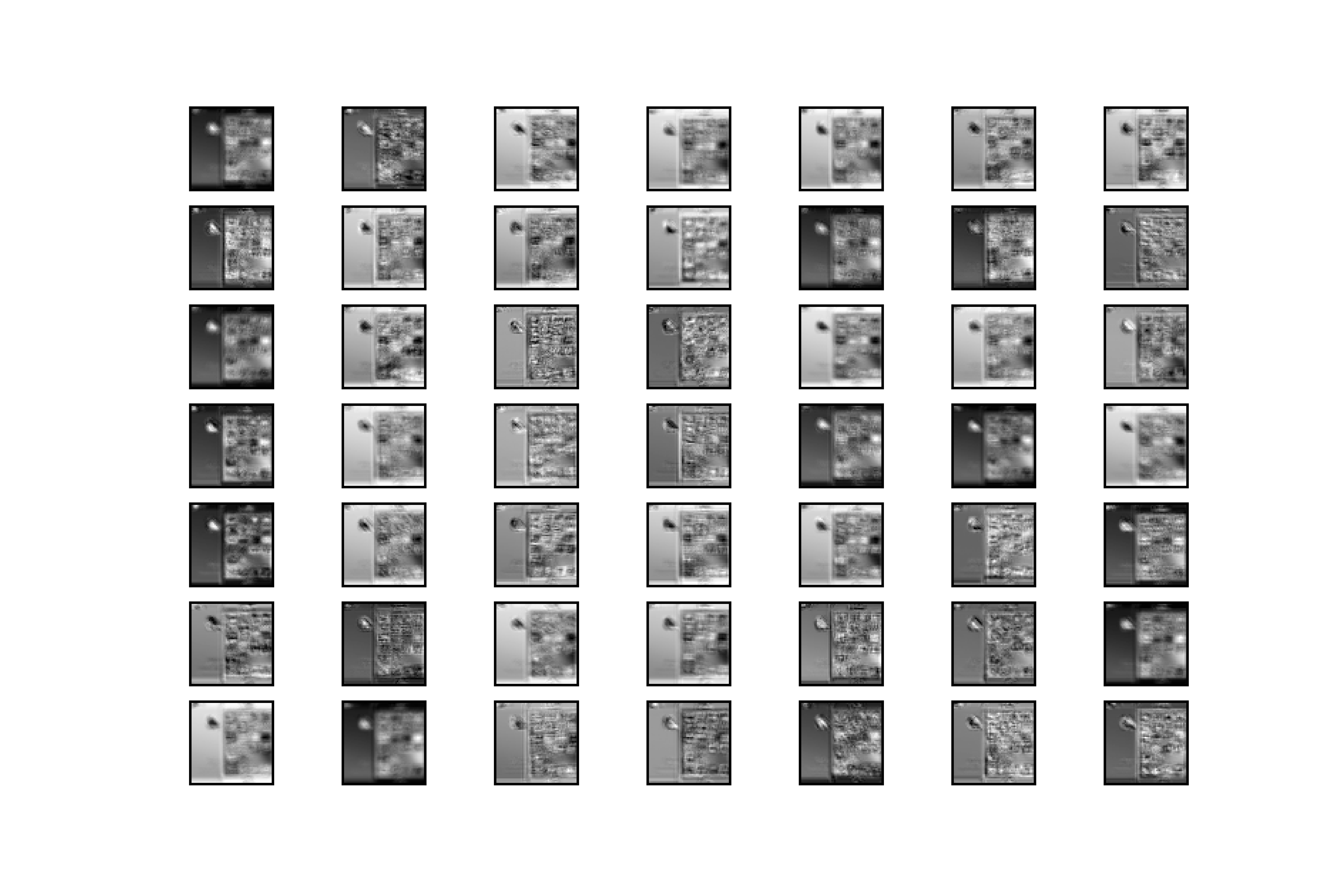}}}%
    \hfill
    \subfloat[Third layer of convolution.]{{\includegraphics[width=0.3\linewidth]{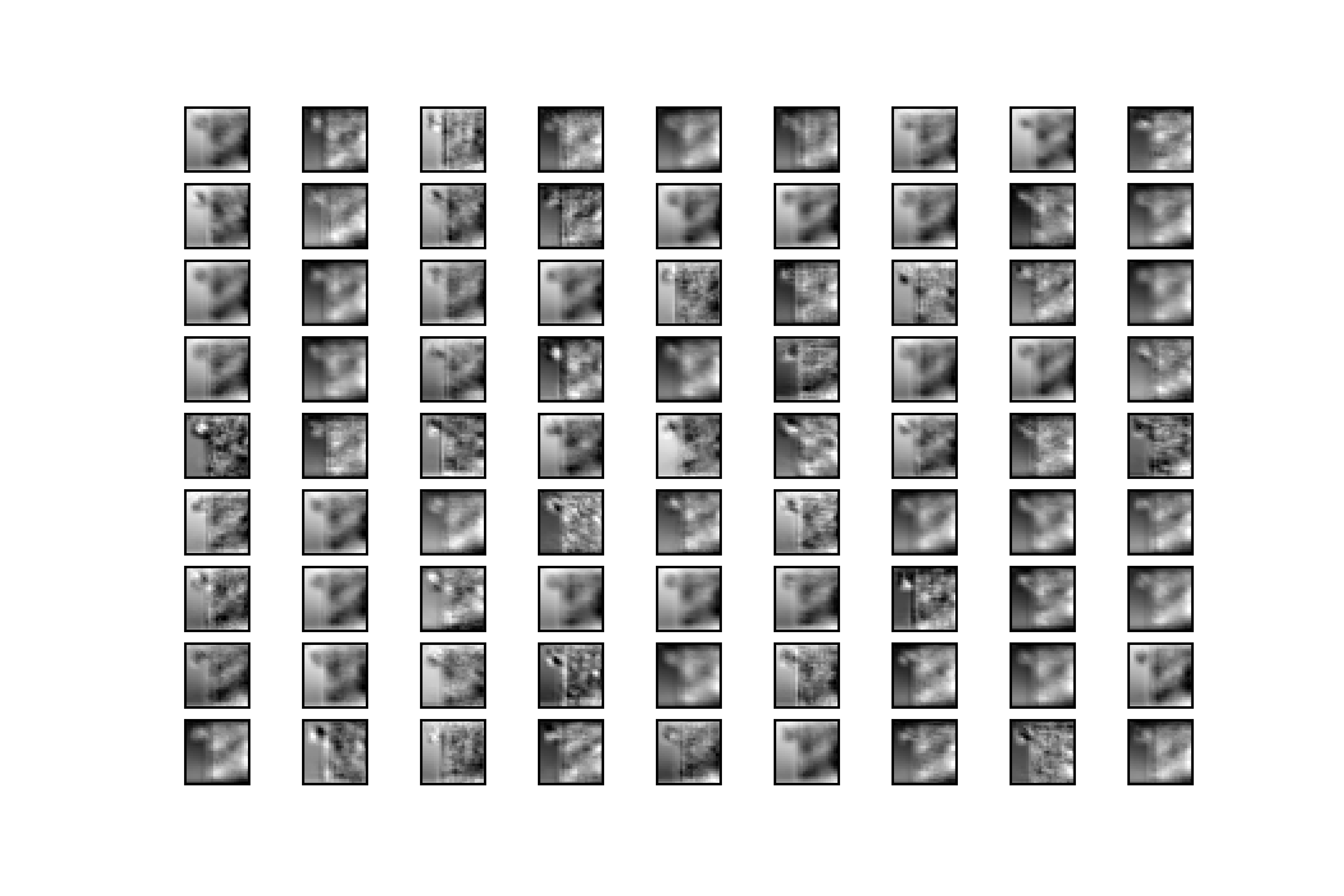}}}%
    \hfill
    \\
    \subfloat[First maxpooling.]{{\includegraphics[width=0.3\linewidth]{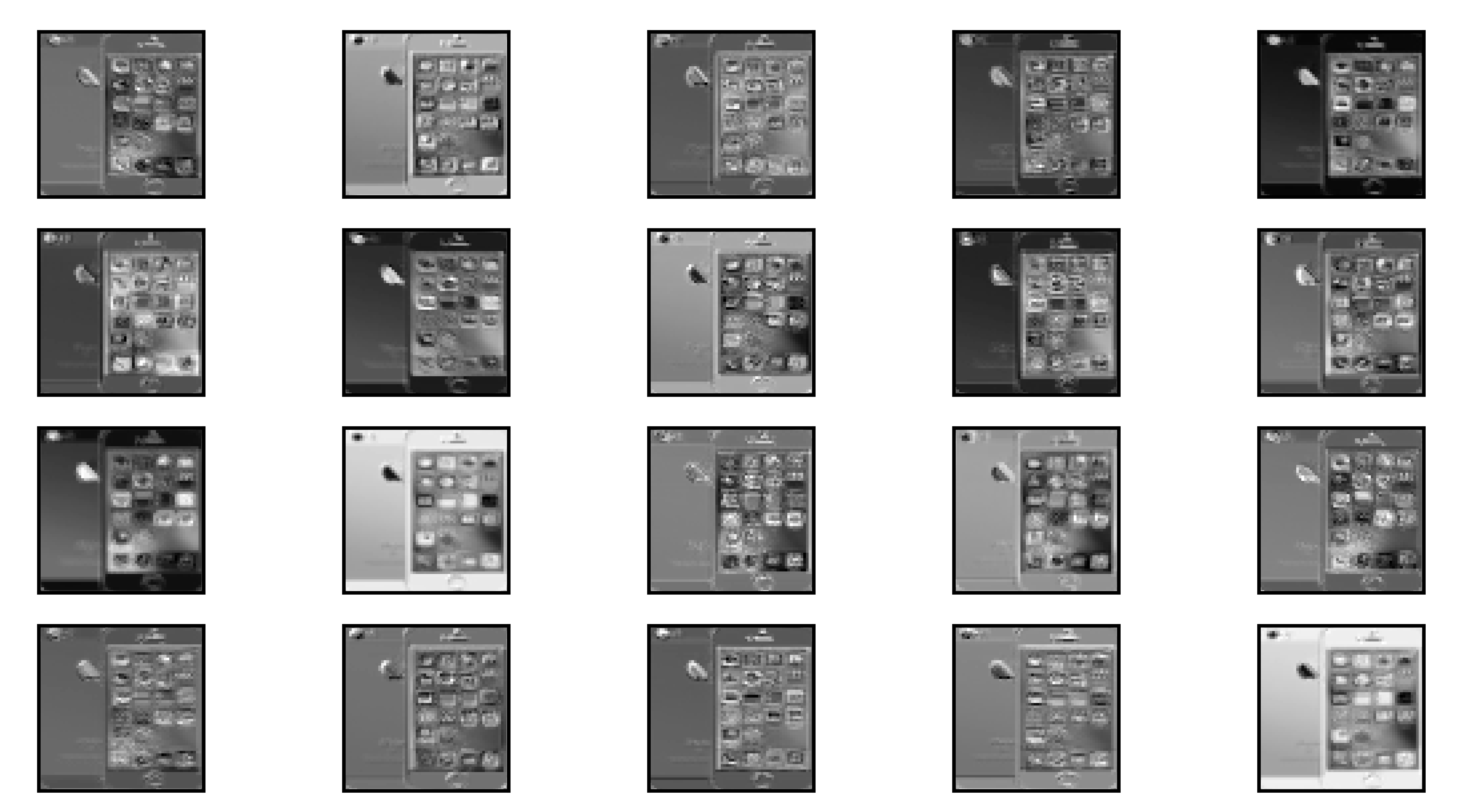}}}%
    \hfill
    \subfloat[Second maxpooling.]{{\includegraphics[width=0.3\linewidth]{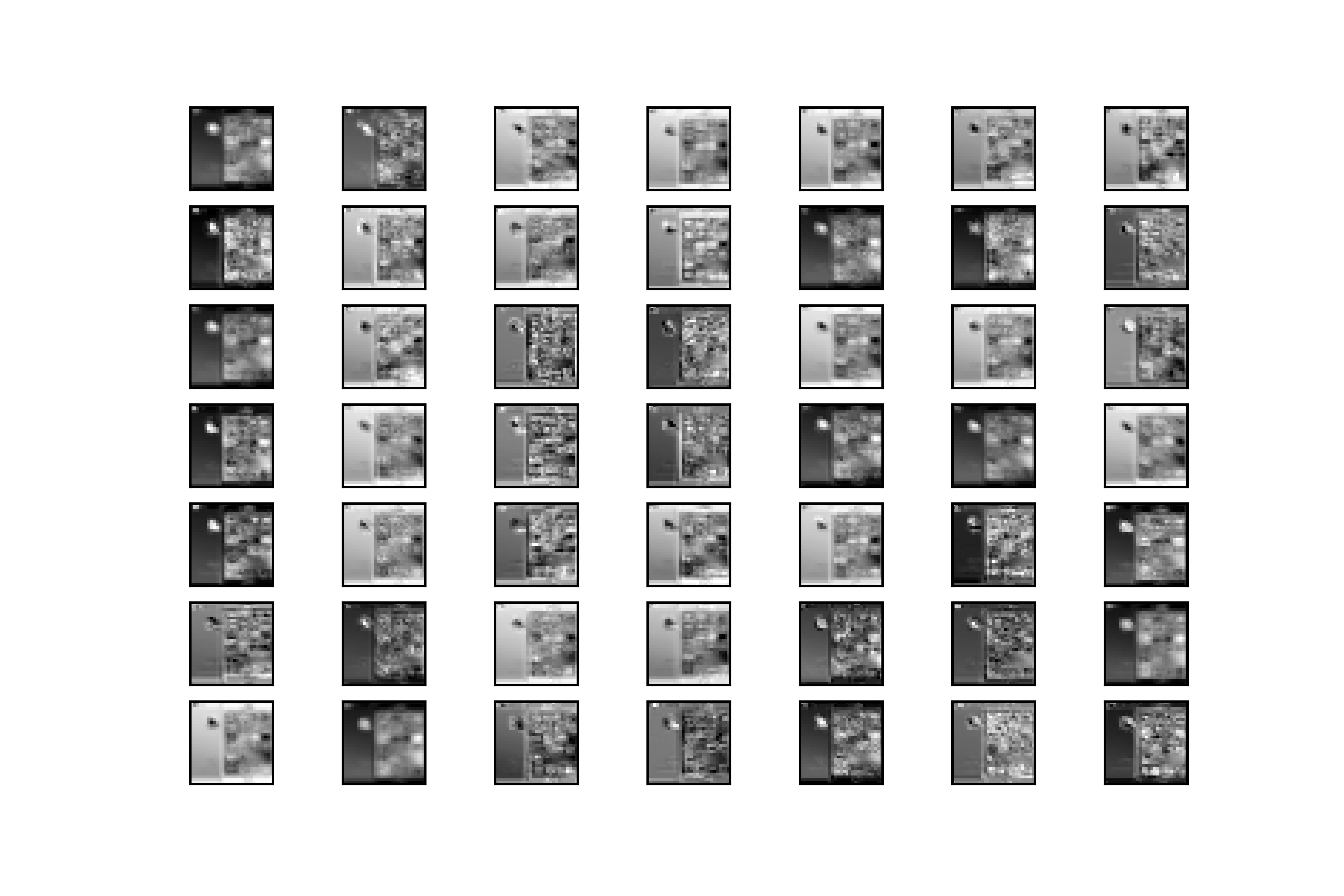}}}%
    \hfill
    \subfloat[Third maxpooling.]{{\includegraphics[width=0.3\linewidth]{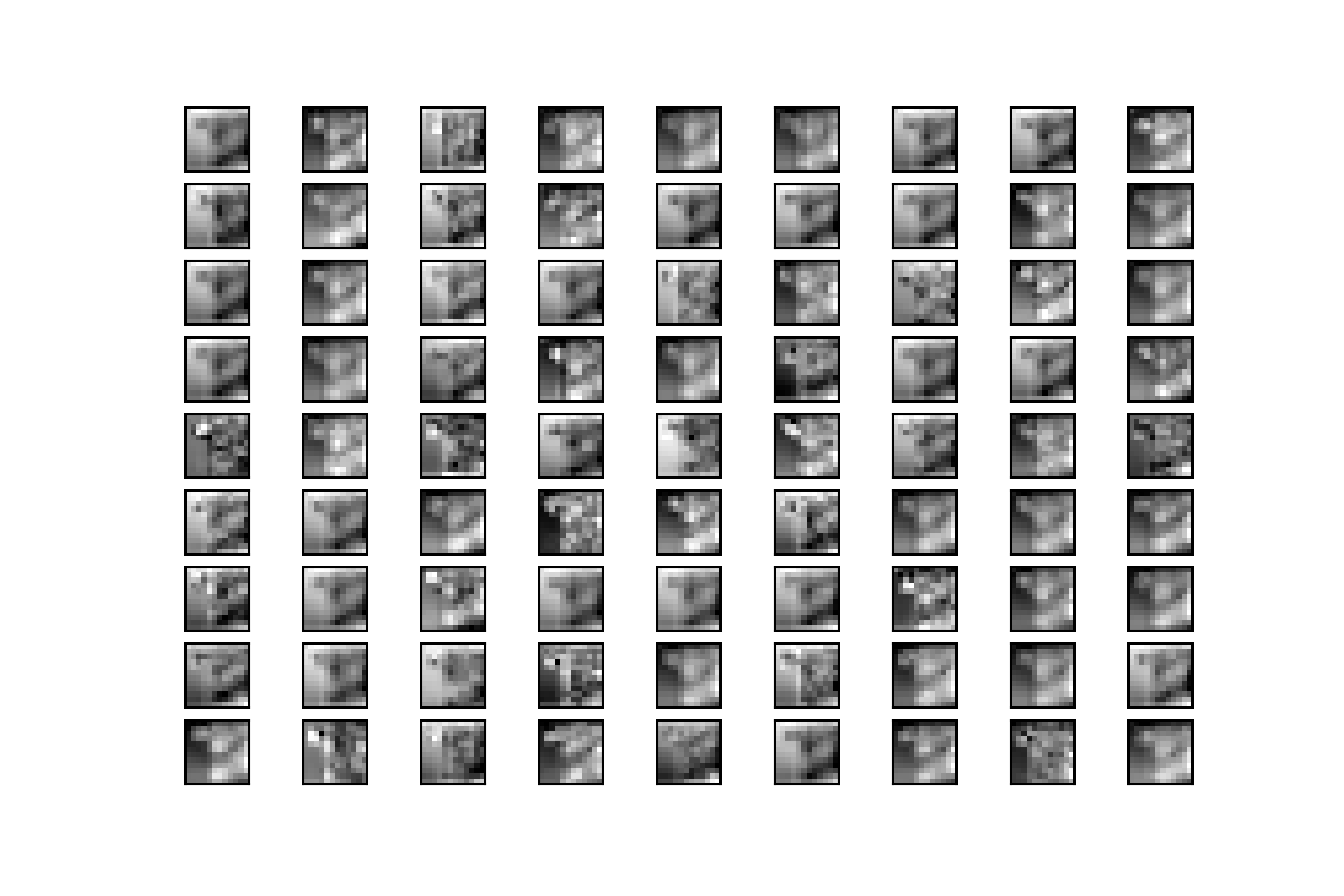}}}%
    \caption{Visualization of different convolutional layers for model 3.}
    \label{Fig:CNN_otput_Model_3}
\end{figure*}

To show how each convolutional layer extracts features, a cellphone image has been injected into the input of the CNN model as an example and the outputs of each convolution layer have been represented in Figure \ref{Fig:CNN_otput_Model_3}. It is obvious that in each layer more features are obtained from the input image which can improve the proposed price prediction model.

\subsection{Model 4: Fully convolutional network}\label{Subsec:model_4}
\begin{figure*}
    \centering
    \includegraphics[width=\textwidth]{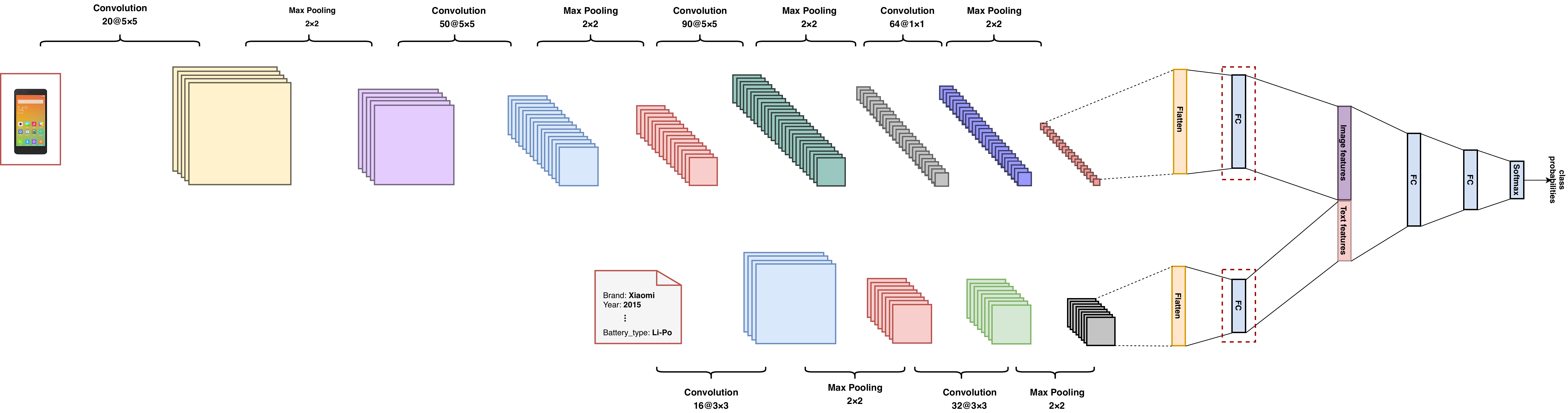}
    \caption{Neural network architecture of method 4. The only difference between method 4 and 5 is on "FC" layers that appear in method five.}
    \label{Fig:model_4}
\end{figure*}

The important part that makes a difference between the fourth and third proposed models is that a convolutional neural network is also used to extract features from non-graphical inputs. As shown in Figure \ref{Fig:model_4}, non-graphical features as a matrix structured format have been fed into the input of the CNN architecture. This neural network includes two convolutional and maxpooling layers. It should also be noted that the image features are obtained just like the third model. Then, the flattened layers appear that pass the extracted features to the concatenation part. In the next step, the concatenation of both graphical and non-graphical is done. Finally, the vector representations of cellphones make the inputs of the classification part.

\subsection{Model 5: Convolutional image and non-graphical features with dense concatenating}\label{Subsec:model_5}
The main goal of this method is to have a higher level feature extraction from non-graphical and graphical inputs. Hence, a dense layer on the output of the convolutional networks is added in extracting features from both non-graphical and image inputs.  As Figure \ref{Fig:model_4} shows, two and three convolution layers calculate non-graphical and image features, respectively. After flattening the extracted embeddings from the convolution layers, they are fed into the fully connected layer. Then, we have the concatenation and classification parts, accordingly, which are the same as the previous proposed models.

\section{Experiment} \label{Sec:Experiments}
In this section, the applied datasets for the proposed methods are discribed. Then, the parameter settings are explained. Finally, the metrics used to evaluate the proposed algorithms are defined.

\subsection{Dataset} \label{Subsuc:Dataset}
The different proposed methods are evaluated on two different datasets Sharma and CD18. In the following paragraphs, these datasets are introduced:

\subsubsection*{\textbf{Sharma}:} Sharma \cite{Sharma}  is a dataset of cellphone crawled from GSMArena\footnote{\url{www.gsmarena.com}} in 2018. The dataset consists of $2000$ row cellphone data with $21$ non-graphical features. In this dataset all features are numerical and  the prices fall into four categories defined as phones' labels.

\subsubsection*{\textbf{CD18}:} GSMArena is a specialized and reputable website that provides cellphones' specifications. On this website, each cellphone has a page that consists of its specifications.
Sharma dataset contains non-graphical information about cellphones. Since, in this study four multimodal approaches have been proposed that need both non-graphical and image information related to the cellphones, it is attempted to collect a dataset called \textit{CD18}\footnote{Cellphone Dataset with 18 Features} from GSMArena to satisfy the need for multimodal dataset. As a mobile phone device has many non-graphical features that are not necessary most of the time, feature selection considers one essential part of the proposed data-dependent project. To reduce the number of cellphone features to those that are trusted to be most helpful for the proposed models, the features that are offered by GSMArena on the top of each cellphone page, have been selected. Figure \ref{Fig:Overal} shows those features on the top of each mobile page. The gathered dataset is made up of the features of this figure. The selected features are listed as below:

\begin{multicols}{2}
    \begin{enumerate}
        \item Brand
        \item Model
        \item Release Date
        \item Weight
        \item Operation System
        \item Storage
        \item Hit
        \item Hit Count
        \item Display Size
        \item Display Resolution
        \item Camera
        \item Video
        \item Processor
        \item Ram
        \item Battery
        \item Battery Type
        \item Picture
        \item Price (Euro)
    \end{enumerate}
\end{multicols}


The gathered data are raw and should be normalized before any utilization. In this regard, changes have been listed as below:
\begin{itemize}
    \item \textit{Weight}: Non-numerical characters such as "gr" have been deleted, and the values of this column are used as a numerical feature.
    \item \textit{OS}: There are too many customized models or names for this feature. To reduce overplus states, this feature has been configured to consist of $20$ models.
    \item \textit{Display size}: Display size quotes as $width \times height$. In this form, the feature should be processed as categorical. To obtain a numerical representation from this feature, this column has been broken into two columns called "V\_resolution" and "H\_resolution" that show the vertical and horizontal pixels, accordingly.
    \item \textit{Video}:  Video's resolution is represented with term "p" as the abbreviation of pixel which is an integer value. To make this feature a numerical one, the term "p" is deleted.
    \item \textit{Processor}: There are a lot of models of cellphone processor's chipset and this makes too many possibilities for the system as a category feature. Thus, only the brands of processors and  $26$ types of these processors have been selected.
    \item \textit{Ram}: For this feature, all values must be on the same unit, so they have been converted to Mb (MegaByte).
    \item \textit{Hit} \& \textit{Hit Count}: These two features do not affect price prediction but only they maybe show the users perspective. So these were not used as non-graphical feature.
\end{itemize}

Moreover, since the price prediction has been considered as a classification task, it is needed to determine the classes of this task. Furthermore, the eligible classes have been created by classifying the cellphone prices into four groups. These groups are estimated by thresholding prices as below:
\begin{itemize}
    \item 0: $< 250$
    \item 1: $250 \leq$ \& $< 500$
    \item 2: $500 \leq$ \& $< 750$
    \item 3: $\geq 750$
\end{itemize}

At the end, the collected dataset contains $3165$ records with $17$ non-graphical features besides the cellphones' images.

\subsection{Parameter settings}
All tests were run via python on a Core i7-5600U 2.6 GHz processor with 12 GB of RAM and Nvidia 840M graphic processor. The learning rate and batch size have been set to 0.001 and 32, respectively. Also, the optimizer has been optimized with the RMSProp optimizer for all five models. Furthermore, the fully connected classifier consists of three layers that the first and second layers include Relu activation function with 300 and 120 neurons, respectively. Also, L1 regularization was set to $\alpha = 0.1$. Finally, the last layer is softmax function that produces four different classes.

\subsection{Evaluation Metrics}
To assess the quality of the proposed methods on the price prediction task, precision, recall, accuracy, and F1-measure are used as the metrics. The reason for deploying these metrics is treating the price prediction as a classification problem. These metrics are defined as follows:

Precision attempts to show what fraction of the retrieved instances is actually correct. While, recall defines
what fraction of factual positives is recognized accurately. Equations \ref{eq:presicion} and \ref{eq:recall}
show how these two metrics are calculated.

\begin{equation}
    Precision = \frac{TP}{TP + FP}
    \label{eq:presicion}
\end{equation}

\begin{equation}
    Recall = \frac{TP}{TP + FN}
    \label{eq:recall}
\end{equation}
here TP, FP, FN and TN show True Positives, False Positives, False Negatives and True Negatives, respectively. 

There are other metrics that rely on both precision and recall such as F1-measure. F1-measure is the harmonic mean of the precision and recall. This metric is defined as Eq. \ref{eq:f_measure}.   
\begin{equation}
    F1-measure = 2 \times \frac{precision \times recall}{precision + recall}
    \label{eq:f_measure}
\end{equation}

Accuracy informally is the fraction of predictions of the proposed model. this metric is calculated based on Eq. \ref{eq:accuracy}.

\begin{equation}
    Accuracy = \frac{TP + TN}{TP + TN + FP + FN}
    \label{eq:accuracy}
\end{equation}
\section{Results} \label{Subsec:Evaluation}
In this section, the proposed approaches are evaluated and compared with three approaches Nasser et al. \cite{Nasser2019}, Subhiksha et al. \cite{Subhiksha2020} and  Asim et al. \cite{Asim2018} that are unimodal and described in section \ref{Sec:Related-study}. Tables \ref{tbl:unimodal_evaluation} and \ref{tbl:multimodal_evaluation} report the obtained results.

Table \ref{tbl:unimodal_evaluation} presents the evaluation of different unimodal methods on two datasets CD18 and Sharma. Just non-graphical features of CD18 have been employed for making a comparison of the proposed unimodal approach with others. It can also be concluded that the proposed unimodal method achieved the best results on both datasets over different metrics. Also, the proposed unimodal approach is the winner of the competition against others in terms of accuracy metric.

Table \ref{tbl:multimodal_evaluation} illustrates the evaluation of the proposed multimodal methods on CD18 dataset using different classifiers. To assess the multimodal methods, six various classifiers consisting SVM \cite{Shi2011}, Logistic Regression, Decision Tree, KNN, and fully connected neural network have been deployed. It can be seen that the classifier constructed based on the deep neural network achieved the best values in each metric, but not in very high deference with compared to other classifiers. Multimodal approaches increase the value of F1-measure and accuracy more than $2\%$.  
Although, using different classifiers have direct effects on the quality of the prediction, it can be seen that in the most cases the concatenation of non-graphical and graphical features (multimodal approach) performs better and makes predictions more accurate than unimodal. The other deduction from the evaluations is that convolution networks can detect features from images of cellphones better than Inception-V3, because cellphones' schema is too close and Inception-V3 is trained to appear different types of objects. In conclusion, due to the results, it can be asserted that images of products are impressive in predicting price along with other features and the multimodal approaches make predictions more authentic.

\begin{figure}[ht]
\centering
    \subfloat[The second proposed model]{{\includegraphics[width=0.45\linewidth]{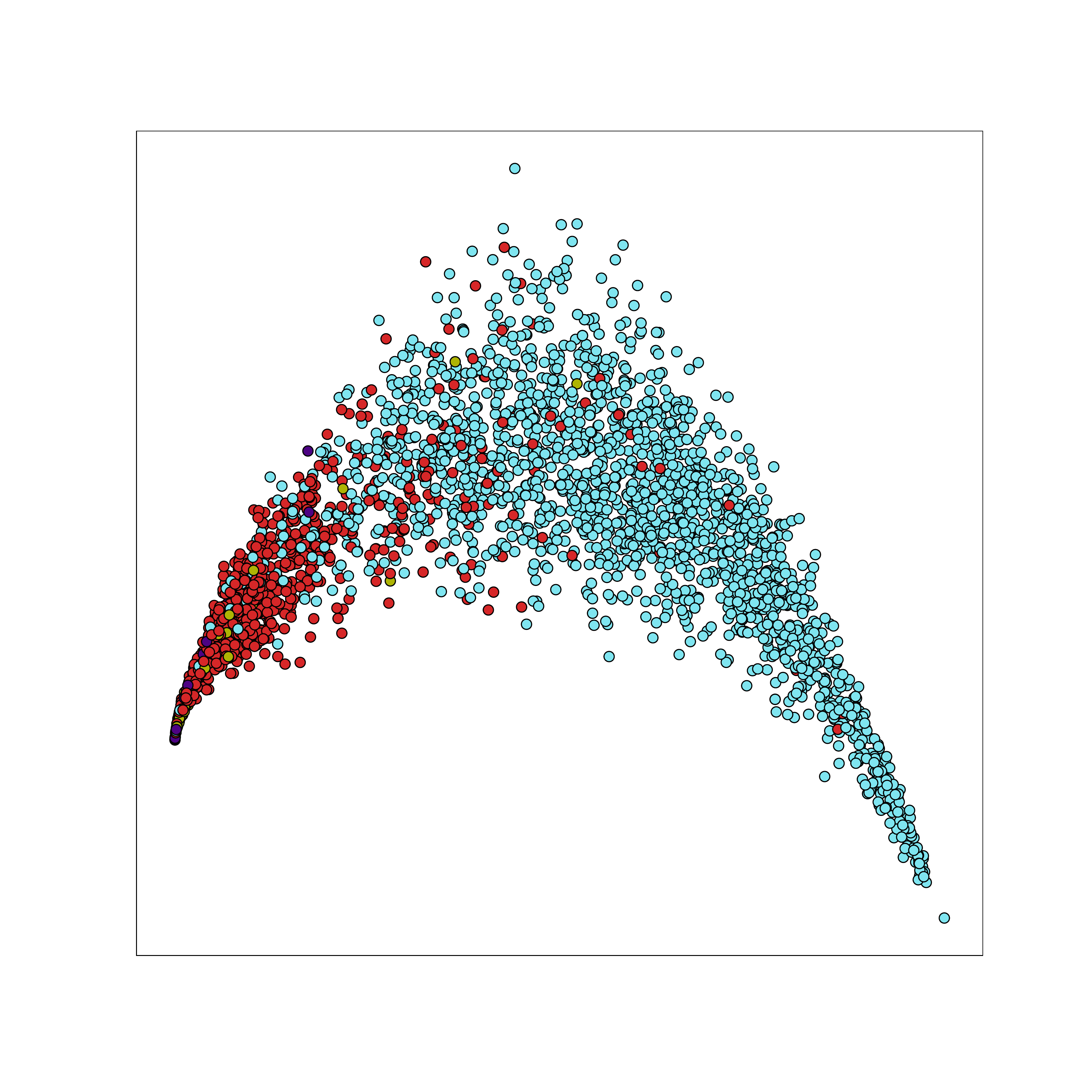}}}%
    \hfill
    \subfloat[The fifth proposed model]{{\includegraphics[width=0.45\linewidth]{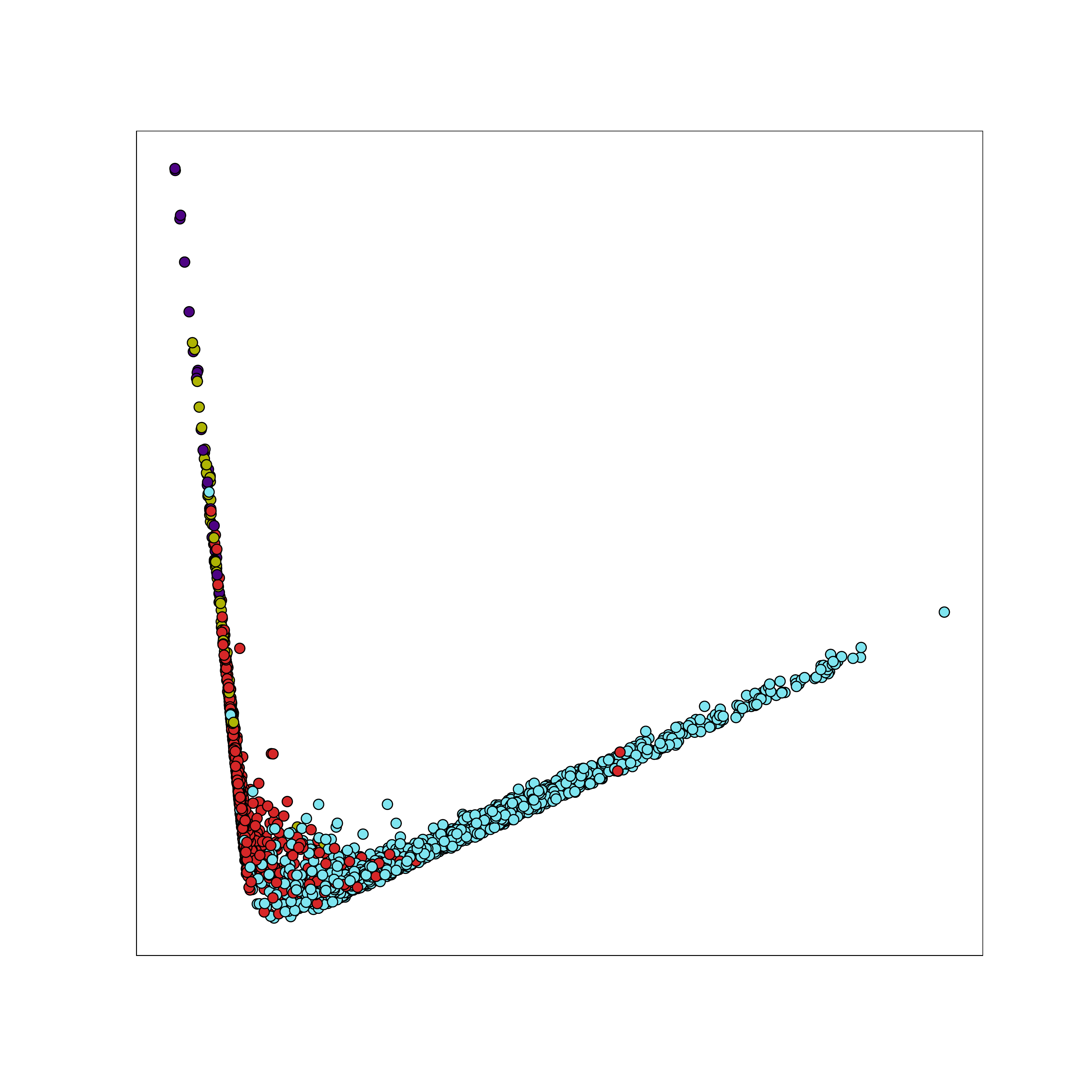} }}%
    \caption{Visualizations of the second and fourth proposed models using PCA. Each point corresponds to a cellphone's price class.
}%
    \label{Fig:visualization}%
\end{figure}

Each cellphone has been presented by a low-dimensional vector that allows us to visualize cellphones
to understand their price ranges. Visualizations of the second and fourth proposed models are shown in Figure \ref{Fig:visualization}. Since the underlying community is known, the community label is used to color the cellphone. It can be observed that embeddings generated by the fourth proposed multimodal approach can separate the communities of mobiles based on their predicted prices better than the second model.

\begin{table}[]
    \centering
    \caption{The results of the proposed unimodal method in comparison with state-of-the-art methods in unimodal datasets. Boldface values indicate the best in metrics.}
    \label{tbl:unimodal_evaluation}
    \begin{tabular}{c|cccc|cccc}
        &\multicolumn{8}{c}{Dataset}    \\
        \cline{2-9}
        \multirow{2}{*}{Method} &\multicolumn{4}{c|}{Sharma \cite{Sharma}}   &\multicolumn{4}{c}{Unimodal dataset of CD18}   \\
        &Precision  &Recall  &Accuracy  &F-measure  &Precision  &Recall &Accuracy   &F-measure  \\
        \hline
        \hyperref[Subsec:model_1]{Proposed Method 1} &\textbf{83.0}   &\textbf{83.0}   &\textbf{82.9}   &\textbf{82.5}   &\textbf{88.0}  &\textbf{85.5}   &\textbf{85.5}  &\textbf{86.1}   \\
        Nasser et al. \cite{Nasser2019} &47.5   &25.4   &25.6   &33.1   &45.8   &67.6   &67.7   &54.6   \\
        Subhiksha et al. \cite{Subhiksha2020}   &   &   &81.0   &   &   &   &   &   \\
        Asim et al. \cite{Asim2018} &   &   &78.0 &   &   &   &   & \\
        \hline
    \end{tabular}
\end{table}

\begin{table}[]
    \centering
    \caption{The results of the proposed multimodal methods in CD18 with different classifier of the last classification part. Boldface values indicate the best value in each metric.}
    \label{tbl:multimodal_evaluation}
    \begin{tabular}{c|cccc|cccc|cccc}
        \multirow{2}{*}{Method} &\multicolumn{4}{c|}{Logistic Regression} &\multicolumn{4}{c|}{KNN}   &\multicolumn{4}{c}{Decision Tree} \\
        &\rotatebox{90}{Precision}  &\rotatebox{90}{Recall} &\rotatebox{90}{Accuracy}   &\rotatebox{90}{F-measure}  &\rotatebox{90}{Precision}  &\rotatebox{90}{Recall} &\rotatebox{90}{Accuracy}   &\rotatebox{90}{F-measure}  &\rotatebox{90}{Precision}  &\rotatebox{90}{Recall} &\rotatebox{90}{Accuracy}   &\rotatebox{90}{F-measure}   \\
        \hline
        \hyperref[Subsec:model_2]{Proposed Method 2}    &83.6   &81.7   &87.5   &82.3   &89.3   &\textbf{88.0}   &87.7  &88.2   &83.4   &83.0   &87.5   &83.2    \\
        \hyperref[Subsec:model_3]{Proposed Method 3}    &88.3   &87.4   &87.4   &87.6   &84.7   &84.2   &84.3   &84.4   &83.4   &81.8   &81.8   &82.1   \\
        \hyperref[Subsec:model_4]{Proposed Method 4}  &80.3   &79.9   &79.8   &79.8   &81.0   &73.6   &73.6   &76.8   &71.3   &69.8   &69.8   &70.5     \\
        \hyperref[Subsec:model_5]{Proposed Method 5}  &85.2   &82.3   &82.4   &83.8   &82.1   &79.9   &79.8   &81.0   &74.9   &74.2   &74.2   &74.5 
        
          \\
        \hline
    \end{tabular}
\end{table}

\begin{table}[]
    \centering
    \label{tbl:multimodal_evaluation_2}
    \begin{tabular}{c|cccc|cccc|cccc}
        \multirow{2}{*}{Method}  &\multicolumn{4}{c|}{SVM}   &\multicolumn{4}{c|}{Gradient Boosting} &\multicolumn{4}{c}{Fully connected}   \\
        &\rotatebox{90}{Precision}  &\rotatebox{90}{Recall} &\rotatebox{90}{Accuracy}   &\rotatebox{90}{F-measure}  &\rotatebox{90}{Precision}  &\rotatebox{90}{Recall} &\rotatebox{90}{Accuracy}   &\rotatebox{90}{F-measure}  &\rotatebox{90}{Precision}  &\rotatebox{90}{Recall} &\rotatebox{90}{Accuracy}   &\rotatebox{90}{F-measure}  \\
        \hline
        \hyperref[Subsec:model_2]{Proposed Method 2}    &88.1   &86.8   &87.9   &87.3   &85.6   &84.9   &87.9   &85.2   &\textbf{89.4}  &86.8  &86.8  &87.8    \\
        \hyperref[Subsec:model_3]{Proposed Method 3}    &89.1   &87.9   &88.0   &\textbf{88.3}   &87.2   &84.9   &84.9   &85.9   &89.1  &\textbf{88.0}  &\textbf{88.0}  &\textbf{88.3}    \\
        \hyperref[Subsec:model_4]{Proposed Method 4}   &83.9   &81.1   &81.2   &82.4   &73.0   &71.7   &71.7   &72.2   &83.7  &83.0  &83.0  &82.9      \\
        \hyperref[Subsec:model_5]{Proposed Method 5}    &85.2   &82.4   &82.3   &83.7   &81.2   &79.8   &79.8   &80.5   &88.1  &83.0  &83.0  &85.3 \\
        \hline
    \end{tabular}
\end{table}

\section{Conclusion} \label{Sec:conclusion}
Proposing a multimodal price prediction system is the primary goal of this paper. The first step of this study was to gather an appropriate dataset, called CD18, that consists of both graphical and non-graphical features of cellphones with their prices. To achieve this goal, information related to the cellphones including their specifications and pictures were collected from GSMArena. After that, five methods were proposed that one of them is unimodal and the rest of them are multimodal approaches. The multimodal methods were created by Inception-V3 and convolutional neural networks that are applied to the graphical, and both graphical and non-graphical features, respectively. Then, there are two parts including the joint part that combines  graphical and non-graphical features, and the classification part that has a dense neural network. The evaluation of the proposed methods was done on two different datasets Sharma and CD18. The results show that the proposed methods give better performance in comparison with the other available researches.



%
\section*{Declarations}

\subsection*{Funding}
The authors did not receive support from any organization for the submitted work.

\subsection*{Conflict of interest}
The authors declare that they have no conflict of interest.



\bibliographystyle{unsrtnat}
\bibliography{Bibliography.bib}   

\end{document}